%% file: main.tex
\documentclass[conference]{IEEEtran}
\ifCLASSINFOpdf
\else
\fi
\usepackage[utf8]{inputenc} 
\usepackage[spanish,es-tabla]{babel} 
\usepackage{todonotes} 

\usepackage{graphicx}
\usepackage[fleqn]{amsmath}
\usepackage{listings}
\usepackage[hidelinks]{hyperref}
\usepackage{subcaption}
\usepackage{booktabs}
\usepackage{amssymb}
\usepackage{amsmath}
\usepackage[most]{tcolorbox}

\input{./math_preamble}

\definecolor{myblue}{RGB}{0,0,100}
\definecolor{mygray}{RGB}{191,191,191}

\tcbset{
  frame code={}
  center title,
  left=0pt,
  right=0pt,
  top=0pt,
  bottom=0pt,
  colback=mygray,
  colframe=white,
  width=\dimexpr\textwidth\relax,
  enlarge left by=0mm,
  boxsep=5pt,
  arc=0pt,outer arc=0pt,
}

\begin{document}
%
\title{Revisión de Métodos de Planificación de Camino de Cobertura para Entornos Agrícolas}

\author{
    \IEEEauthorblockN{Ismael Ait, Ernesto Kofman y Taihú Pire}
    \IEEEauthorblockA{CIFASIS, Centro Internacional Franco Argentino de Ciencias de la Información y de Sistemas (CONICET-UNR), Argentina}
}

\maketitle

\begin{abstract}
\input{src/abstract.tex}
\end{abstract}

\IEEEpeerreviewmaketitle

\section{Introducción}
\label{sec:introduction}
\input{src/intro.tex}
\section{Trabajo Relacionado}
\label{sec:related}
\input{src/related.tex}

\section{Algoritmos de path planning de cobertura}
\label{sec:method}
\input{src/method.tex}

\section{Discusión: combinación celda + franjas}
\label{sec:proposal}
\input{src/proposal.tex}

\section{Conclusiones y trabajo futuro}
\label{sec:conclusions}
\input{src/conclusions.tex}


\bibliographystyle{IEEEtran}
\bibliography{biblio}

\end{document}

%% file: math_preamble.tex
\usepackage{amsmath}
\usepackage{amssymb}
\usepackage{amsopn}

\newcommand{\cell}[1]{\text{\textcircled{\footnotesize{\textit{#1}}}}}

%% file: src/abstract.tex
 El uso de un método eficiente de planificación de cobertura es clave para la navegación autónoma en entornos agrícolas, en los cuales un robot debe cubrir grandes extensiones de cultivos. 
 En este trabajo se repasa de manera general el estado del arte actual de los métodos de planificación de camino de cobertura. 
 Dos técnicas ampliamente utilizadas y aplicables a entornos agrícolas son descritas en detalle.
 La primera, consiste en la descomposición de un campo complejo con obstáculos en subregiones más simples conocidas como celdas, para posteriormente generar un patrón de cobertura en cada una de ellas.
 La segunda, analiza espacios compuestos por franjas paralelas por las cuales el robot debe circular, con el fin de encontrar el orden de visita de franjas óptimo que minimice la distancia total recorrida.
 Adicionalmente, se discute la combinación de ambas técnicas con el fin de obtener un plan de cobertura global más eficiente. 
 Dicho análisis fue concebido para ser implementado con el robot desmalezador de cultivos de soja desarrollado en el CIFASIS (CONICET-UNR).

%% file: src/intro.tex
 
 En los últimos años el uso de robots autónomos móviles ha experimentado un crecimiento considerable en diferentes sectores de la industria.
 La inevitable necesidad de aumentar la producción de alimentos debido al crecimiento en la población mundial, ha despertado en la industria agrícola un singular interés sobre estas nuevas tecnologías.
 En la actualidad, ya podemos encontrar el empleo de robots móviles en la asistencia o ejecución de diversas tareas en este campo \cite{vougioukas2019agricultural}.
 
 La planificación de caminos \textit{(Path Planning)} es una pieza fundamental para concretar la navegación autónoma de los robots móviles.
 Concretamente, los algoritmos de \textit{Path Planning} buscan determinar un camino libre de obstáculos desde un punto inicial hasta un punto meta, buscando principalmente minimizar la distancia recorrida por el robot.
 
 Una variante de la tarea de \textit{Path Planning}, conocida como Planificación de Camino de Cobertura (CPP por sus siglas en inglés \textit{Coverage Path Planning}), trata la búsqueda de un camino que pase por todos los puntos de un espacio determinado \cite{choset1998coverage}. 
 La planificación de camino de cobertura posee especial utilidad en entornos agrícolas, donde se suele tener la necesidad de recorrer un espacio de cultivos de manera completa, mientras el robot realiza diferentes tareas como la de preparación del terreno, siembra, fertilización, y cosecha entre otras.
 Adicionalmente, los algoritmos de CPP se utilizan en muchas otras aplicaciones de robótica como ser aspiradoras de pisos \cite{decarvalho1997complete}, \cite{dakulovic2011complete}, detectores de minas \cite{acar2003path}, \cite{acar1999sensor}, cortadoras de césped autónomas \cite{yang2004aneural} y limpiadores de ventanas \cite{farsi1994robot}, solo por nombrar algunas.
 
 Existen en la literatura muchos trabajos que abordan el problema de CPP. 
 Sin embargo, no todos los métodos pueden aplicarse en cualquier escenario de cobertura.
 Resultaría inviable que un robot con restricciones en sus movimientos (como por ejemplo un automóvil) intentara ejecutar un plan de cobertura pensado para robots omnidireccionales.
 De manera similar, no sería adecuado usar un método de cobertura aproximado, que no garantice que el robot pase por todos los puntos del espacio, en una aplicación de detección de minas pero sí quizás en una aplicación de aspiradoras de piso.
 Es fundamental tener en consideración las características tanto del espacio a cubrir como del robot que realiza la cobertura para elegir de manera adecuada un método viable y eficaz.
 
 En los entornos agrícolas, una de las características principales es que el robot debe recorrer el campo siguiendo las hileras de cultivos, para que de esta manera evite pisarlos.
 Otra característica relevante, es que los robots agrícolas suelen ser vehículos no holonómicos, es decir, que poseen restricciones en sus movimientos.
 Por esto último, también es importante tener en cuenta la dirección de movimiento y ángulo de giro del robot.
 
\begin{figure}[!htbp]
    \centering
    \includegraphics[width=0.7\columnwidth]{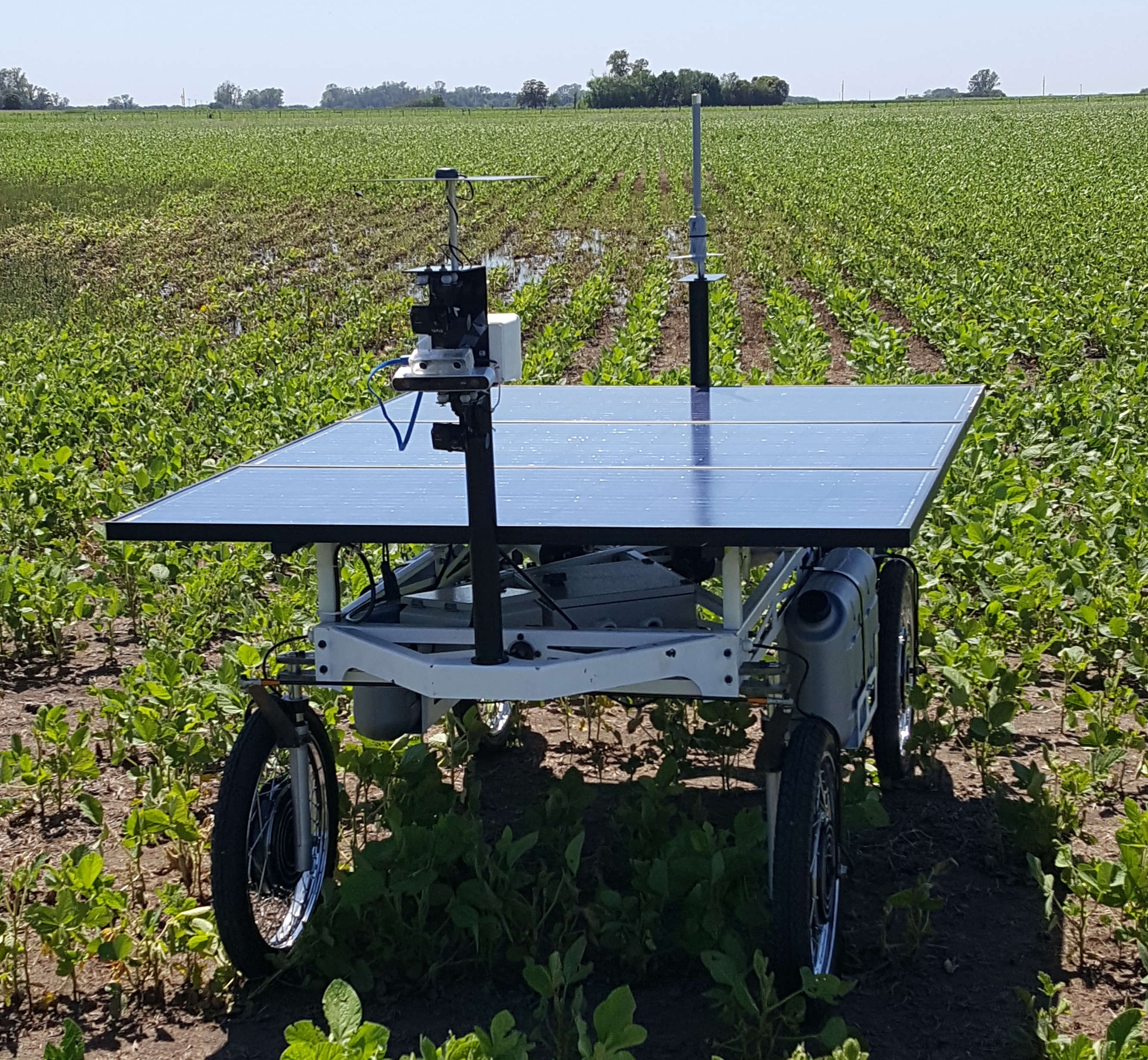} 
    \caption{Robot desmalezador desarrollado en el CIFASIS.}
    \label{fig:weed_removing_robot}
\end{figure}

 En este trabajo, realizamos una revisión del estado del arte de los algoritmos actuales de CPP, poniendo especial foco en aquellos de mayor utilidad en entornos agrícolas. 
 Adicionalmente, proponemos una combinación de métodos que permita desarrollar un plan de cobertura de manera eficiente.
 Este trabajo esta enmarcado dentro del desarrollo del robot desmalezador (ver Figura~\ref{fig:weed_removing_robot}), llevado adelante por el CIFASIS, Rosario.
 Dicho robot tiene como objetivo desplazarse a través de los surcos de campos de soja para detectar las malezas entre los cultivos y aplicar agroquímicos de forma localizada, evitando de esta manera dañar el entorno u otras personas \cite{pire2019rosario}.
 
 El resto de este trabajo se encuentra estructurado de la siguiente manera. 
 En la sección \ref{sec:related} se describen los trabajos existentes de CPP en general, como así también los utilizados específicamente en entornos agrícolas. 
 En la sección \ref{sec:method} se presentan dos métodos ampliamente estudiados, que resultan de especial interés en el proyecto del robot desmalezador.
 En la sección \ref{sec:proposal} se discute cómo pueden combinarse ambas técnicas de manera eficiente.
 Por último, en la sección \ref{sec:conclusions} desarrollamos las conclusiones finales y el trabajo a futuro.

%% file: src/related.tex
 La planificación de camino de cobertura es un problema abordado desde hace ya algunas décadas y muchos trabajos fueron realizados en pos de obtener las mejores soluciones para los diferentes entornos de navegación existentes. 
 Trabajos como \cite{choset2001coverage}, \cite{galcernan2013asurvey} y \cite{khan2017oncomplete}, presentan revisiones exhaustivas de los diferentes métodos de cobertura y sus principales aplicaciones para cada uno. 
 Particularmente, en \cite{choset2001coverage} se pone el foco en la planificación de trayectorias de cobertura ejecutadas por un único robot en el plano, mientras que en \cite{galcernan2013asurvey} se analizan también casos de cobertura en tres dimensiones y la ejecución de coberturas por medio de múltiples robots.
 Por otra parte, \cite{khan2017oncomplete} se enfoca en robots no holonómicos, es decir, cuando existen restricciones en los grados de libertad que posee el robot con respecto al espacio a cubrir.

 Los algoritmos de CPP pueden clasificarse en \textit{completos} y \textit{heurísticos} \cite{choset2001coverage}, dependiendo de si se puede o no dar garantías de una cobertura completa del espacio libre. 
 De manera independiente, también se los puede clasificar como algoritmos \textit{off-line} y \textit{on-line} \cite{galcernan2013asurvey}. 
 Los algoritmos \textit{off-line} se basan en información estacionaria, y se asume que el espacio a cubrir es conocido de antemano. 
 En cambio, los algoritmos \textit{on-line} no asumen conocimiento previo del espacio. 
 Por el contrario, usan las mediciones obtenidas de los sensores como información para la generación del plan de cobertura. 

 
 En entornos agrícolas, el espacio a cubrir por el robot generalmente se trata de un campo compuesto por hileras de cultivos paralelas entre sí.
 El conjunto de hileras que el robot puede abarcar en un mismo momento se lo conoce en la literatura como franja (\textit{swath} o \textit{track}).
 Los bordes del campo en los extremos de las hileras donde el robot realiza los giros para posicionarse en la siguiente franja se los conoce como cabeceras del campo (\textit{headlands}).
 En los trabajos \cite{bochtis2008minimising} y \cite{hammed2011driving} se presenta el problema de elegir una secuencia óptima para recorrer las franjas, buscando minimizar la distancia del trayecto no útil que el robot debe realizar en las cabeceras para pasar de una franja a la siguiente.
 
 En \cite{bochtis2008minimising} la optimización se realiza a través de la programación lineal binaria y se demuestra experimentalmente que la secuencia obtenida reduce hasta la mitad la distancia recorrida entre franjas, comparándola con caminos elegidos por operarios experimentados.
 Por otro lado, en \cite{hammed2011driving} la solución se genera por medio de algoritmos genéticos.
 
 Otro trabajo orientado a entornos agrícolas pero que aborda un aspecto un poco diferente es \cite{oksanen2009coverage}.
 En dicho trabajo, se busca la mejor manera de dividir un campo y determinar la dirección de las hileras de cultivos, previo a los trabajos de preparación del terreno y sembrado.

 En el presente trabajo proponemos el uso en conjunto de dos de las técnicas más usadas en la planificación de cobertura en espacios agrícolas.
 La descomposición en celdas del espacio libre de obstáculos y la optimización del orden de recorrido de las franjas.
 Adicionalmente, realizamos un análisis de la importancia de aplicar ambos métodos en conjunto de manera eficiente.

%
%

%% file: src/method.tex
 
 El procedimiento tradicional para obtener un plan de cobertura de espacios complejos consiste en: 
 Primero, descomponer el espacio libre de obstáculos en subregiones más simples denominadas celdas. 
 En segundo lugar, determinar un recorrido que visite cada una de las celdas. 
 Por último, generar un patrón de cobertura para cada celda de manera individual.

 En términos generales, los campos agrícolas están compuestos por hileras de cultivos paralelas entre sí.
 El conjunto de hileras contiguas que el robot puede cubrir a la vez forma lo que denominamos una franja.
 El patrón de cobertura generado debe lograrse de tal forma que el robot circule por las franjas evitando así pisar los cultivos con sus ruedas.
 Al finalizar el recorrido de una franja se debe elegir cuál es la siguiente a cubrir.

 En las subsecciones ~\ref{sec:cell_decomposition} y ~\ref{sec:track_selection} se detallan las técnicas de descomposición en celdas y selección eficiente de franjas respectivamente.
%
\subsection{Descomposición en celdas}
\label{sec:cell_decomposition}
 El método de descomposición en celdas divide un espacio libre de obstáculos, posiblemente complejo, en regiones no superpuestas conocidas como celdas. 
 Con esto se busca obtener espacios que resulten más ``fáciles'' de cubrir usando movimientos simples.
 Un grafo de adyacencia se utiliza para representar esta descomposición, donde cada nodo representa una celda y cada arista representa la adyacencia entre dos celdas contiguas. 
 Por último, se calcula un recorrido exhaustivo por los nodos del grafo para completar la cobertura de todas las celdas.

 En su forma más simple, la descomposición es generada mediante una línea vertical barriendo el espacio de izquierda a derecha. 
 Los límites de las celdas se establecen cuando algún evento ocurre sobre el barrido.

 La generalización matemática de este método simple, se basa en el uso de funciones de Morse \cite{milnor1969morse} y sus puntos críticos, donde se ubican los límites de las celdas. 
 De manera preliminar pensemos en la función $f : \mathbb{R}^2 \rightarrow \mathbb{R}, f(x,y) = x$.
 Cada línea vertical de barrido puede verse como la preimagen de $f$ para un $x \in \mathbb{R}$ fijo. 
 El barrido puede realizarse entonces recorriendo en forma incremental los valores de $x$. 
 
 De manera más general, a esta ``línea'' de barrido se la conoce como corte (\textit{slice}) del espacio y a $f$ como la función de corte.
 A medida que el corte barre el espacio libre, los obstáculos interactúan con éste, tanto sea dividiendo el corte en segmentos más pequeños, como así también fusionando segmentos de menor tamaño en uno mayor.
 Modificando la función de corte se puede cambiar el patrón por medio del cual el robot va a realizar su cobertura.
 
 En \cite{acar2002morse} se presentan varios ejemplos de funciones de corte distintos que resultan útiles para diferentes escenarios de cobertura.
 Para los ambientes agrícolas, el corte debe seguir la dirección de las hileras de cultivos.

 Formalmente, definimos un corte como el conjunto de puntos en el plano que forman la preimagen para un valor fijo $\lambda \in \mathbb{R}$ de una función de corte.
\begin{equation*}
    \mathcal{C}_{\lambda} = \{(x,y) \in \mathbb{R}^2 \: | \: f(x,y) = \lambda\},\; \forall \lambda \in \mathbb{R}.
\end{equation*}

 Denotamos con $\mathcal{L}$ al espacio libre de obstáculos a cubrir y con $\mathcal{O}_{i}$ para $i \in \{1,\dots,n\}$ a los $n$ obstáculos. 

%

 Notamos con $\partial\mathcal{L}$ a los bordes del espacio libre, y de manera similar, con $\partial\mathcal{O}_{i}$ a los bordes del obstáculo $i$.
 Tenemos que $\partial\mathcal{O}_i \subset \partial\mathcal{L}$ para $i \in \{1,\dots,n\}$.

\begin{figure}[!htbp]
    \centering
    \includegraphics[width=0.75\columnwidth]{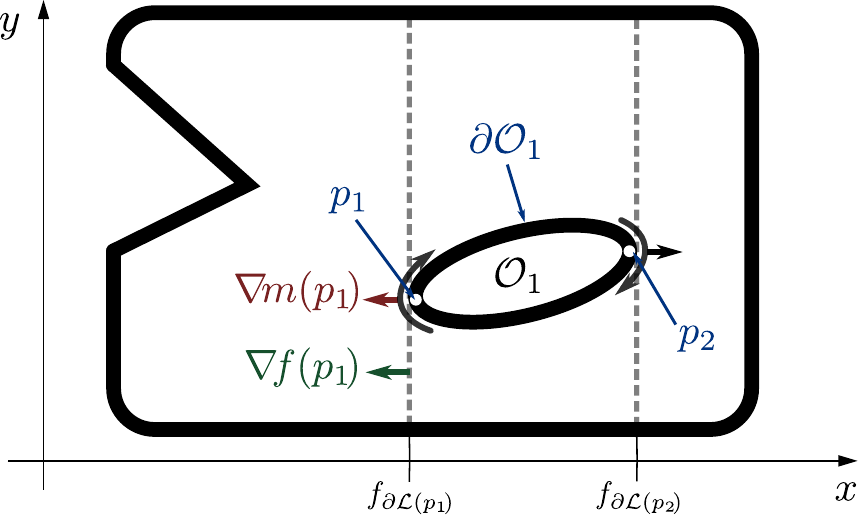} 
    \caption{Puntos críticos $p_1$ y $p_2$ en bordes suaves.}
    \label{fig:smooth}
\end{figure}

 Para definir las celdas, nuestro interés es encontrar los puntos críticos de $f_{\partial\mathcal{L}}$, la restricción de la función $f$ sobre los bordes del espacio libre.
 La Figura~\ref{fig:smooth} muestra dos puntos en los bordes de un obstáculo, $p_1, p_2 \in \partial\mathcal{O}_1$.
 Veremos que estos puntos resultan puntos críticos, es decir, que $f_{\partial\mathcal{L}}$ alcanza extremos locales en $p_1$ y $p_2$.

 La función $f(x,y) = x$ puede ser interpretada como medida de la distancia entre un punto $p$ en el plano y el eje de coordenadas $y$. Del mismo modo, $f_{\partial\mathcal{L}}$ mide la distancia de los puntos que están sobre el obstáculo y el eje de coordenadas $y$. 
 Consideremos ahora los puntos $p \in \partial\mathcal{O}_1$ en el camino que recorre el obstáculo señalado con la flecha de la izquierda en la Figura~\ref{fig:smooth}.
 A medida que nos acercamos a $p_1$ disminuye el valor de $f_{\partial\mathcal{L}}(p)$ y, luego de pasar por $p_1$, el valor de $f_{\partial\mathcal{L}}(p)$ se incrementa.
 En otras palabras, $f_{\partial\mathcal{L}}(p_1)$ es un mínimo local.
 De manera inversa, $f_{\partial\mathcal{L}}(p_2)$ es un máximo local.
 Si asumimos que $\partial\mathcal{O}_i$ puede describirse como la preimagen de una función continua $m_i : \mathbb{R}^2 \rightarrow \mathbb{R}$, es decir, 
\begin{equation*}
\partial\mathcal{O}_i = \{(x,y) \in \mathbb{R}^2 \:|\: m_i(x,y) = 0\},
\end{equation*}
 entonces su gradiente $\nabla m_i(p)$ es el vector normal al perímetro de $O_i$ en el punto $p$.
 Por lo tanto, un punto $p \in \partial\mathcal{O}_i$ es un punto crítico de $f_{\partial\mathcal{L}}$ si y solo si los vectores gradientes de $f$ y $m_i$ en $p$ son paralelos, es decir, $\nabla f (p) \parallel \nabla m_i(p)$.

 Para poder aplicar este método, las funciones de corte restringidas a los bordes de los obstáculos deben ser funciones suaves.
 El método no puede aplicarse directamente si los obstáculos poseen bordes en ángulo.
 El problema con este tipo de obstáculos es que los vectores normales no están definidos en esos puntos.
 Sin embargo, podemos usar la generalización del gradiente de \cite{clarke1990optimization} que define vectores normales a los bordes en puntos no suaves.
 
 El gradiente generalizado es el conjunto de vectores dentro de la envolvente convexa (\textit{convex hull}) de $\nabla m^1(p)$ y $\nabla m^2(p)$, tales que $m^1$ y $m^2$ sean dos funciones en las adyacencias suaves del borde del obstáculo.
 De esta manera, en los puntos no derivables de los bordes, si el gradiente de la función de corte $\nabla f(p)$ (o su vector opuesto $-\nabla f(p)$) está contenido dentro de la envolvente convexa del gradiente generalizado, tendremos un punto crítico.
 Notar que en los puntos suaves, el gradiente generalizado se reduce al gradiente convencional.
 
\begin{figure}[!htbp]
    \centering
    \includegraphics[width=0.75\columnwidth]{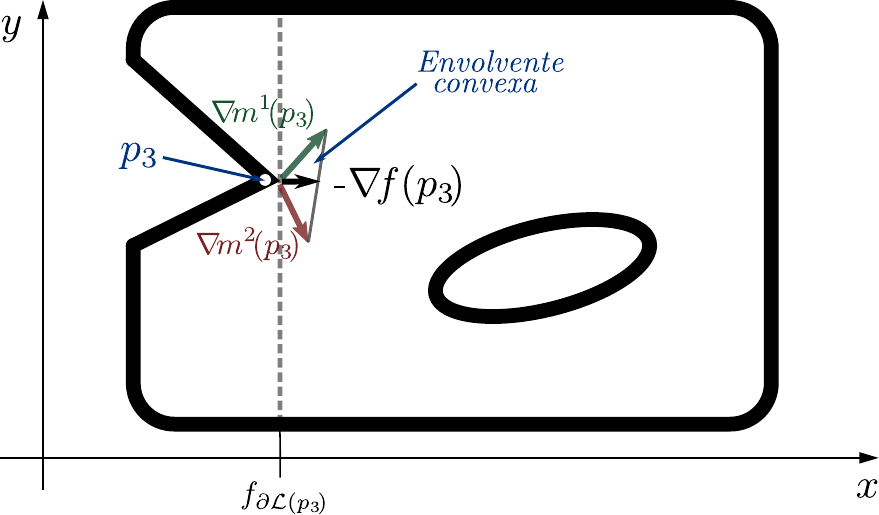} 
    \caption{Punto crítico $p_3$ y su gradiente generalizado.}
    \label{fig:non-smooth}
\end{figure}

 En el espacio de ejemplo, el borde que delimita el campo a cubrir posee puntos no suaves.
 La Figura~\ref{fig:non-smooth} muestra el punto crítico $p_3$, cuya envolvente convexa contiene el vector opuesto al vector normal $\nabla f(p_3)$.

 Una vez concluida la descomposición en celdas, restan dos pasos más para completar el plan de cobertura. 
 Primero se debe determinar un camino exhaustivo que pase por todos los nodos del grafo de adyacencias formado por las celdas, y luego se debe generar un patrón de cobertura para cada celda.
 
 Los patrones de cobertura pueden elaborarse alternando dos movimientos: un movimiento a lo largo del corte y un movimiento a lo largo del borde del obstáculo. 
 El último, dirige al robot hacia el siguiente corte que debe recorrer desplazándolo un ancho del robot sobre el borde del obstáculo. 
 Para la función de corte de ejemplo ${f(x,y) = x}$, el patrón de cobertura resulta similar a un zigzag.
\begin{figure}[!htbp]
    \centering
    \includegraphics[width=0.55\columnwidth]{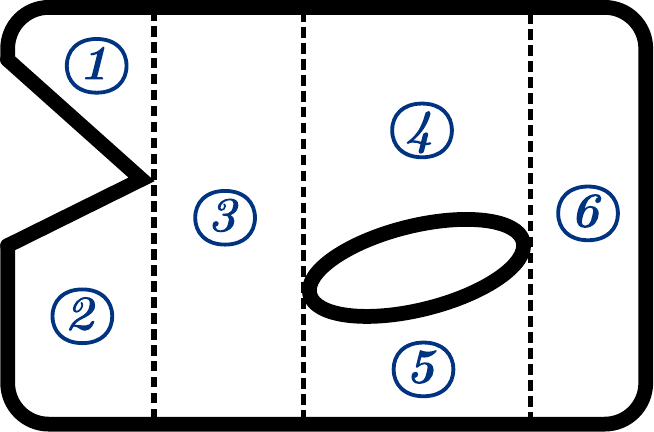}
    \caption{Descomposición en celdas.}
    \label{fig:decomposition}
\end{figure}
 La Figura~\ref{fig:decomposition} exhibe la descomposición en celdas final del espacio libre de obstáculos del ejemplo utilizado.

\subsection{Elección de franjas}
\label{sec:track_selection}
 En entornos agrícolas, el campo a cubrir generalmente está formado por hileras paralelas de cultivos que el robot debe recorrerlas por completo para realizar su tarea encomendada. 
 El patrón a utilizar debe generar recorridos en la misma dirección que estas hileras.
 Dependiendo del ancho del robot y del ancho de los cultivos, cada recorrido que realiza el robot a lo largo del campo agrícola abarca un determinado número de hileras, lo que determina una franja de cobertura.

 Una vez que el robot inicia el recorrido de una franja, sigue por la misma hasta que termina de recorrerla en su otro extremo.
 Al finalizar el recorrido de una franja, se debe elegir la siguiente a recorrer.
 El método descripto en esta subsección, tiene como objetivo optimizar la secuencia en que se recorren las franjas con el objetivo de minimizar la distancia total no productiva que se invierte en realizar los giros entre franjas en las cabeceras del campo.
\begin{figure}[!htbp]
    \centering
    \includegraphics[width=1\columnwidth]{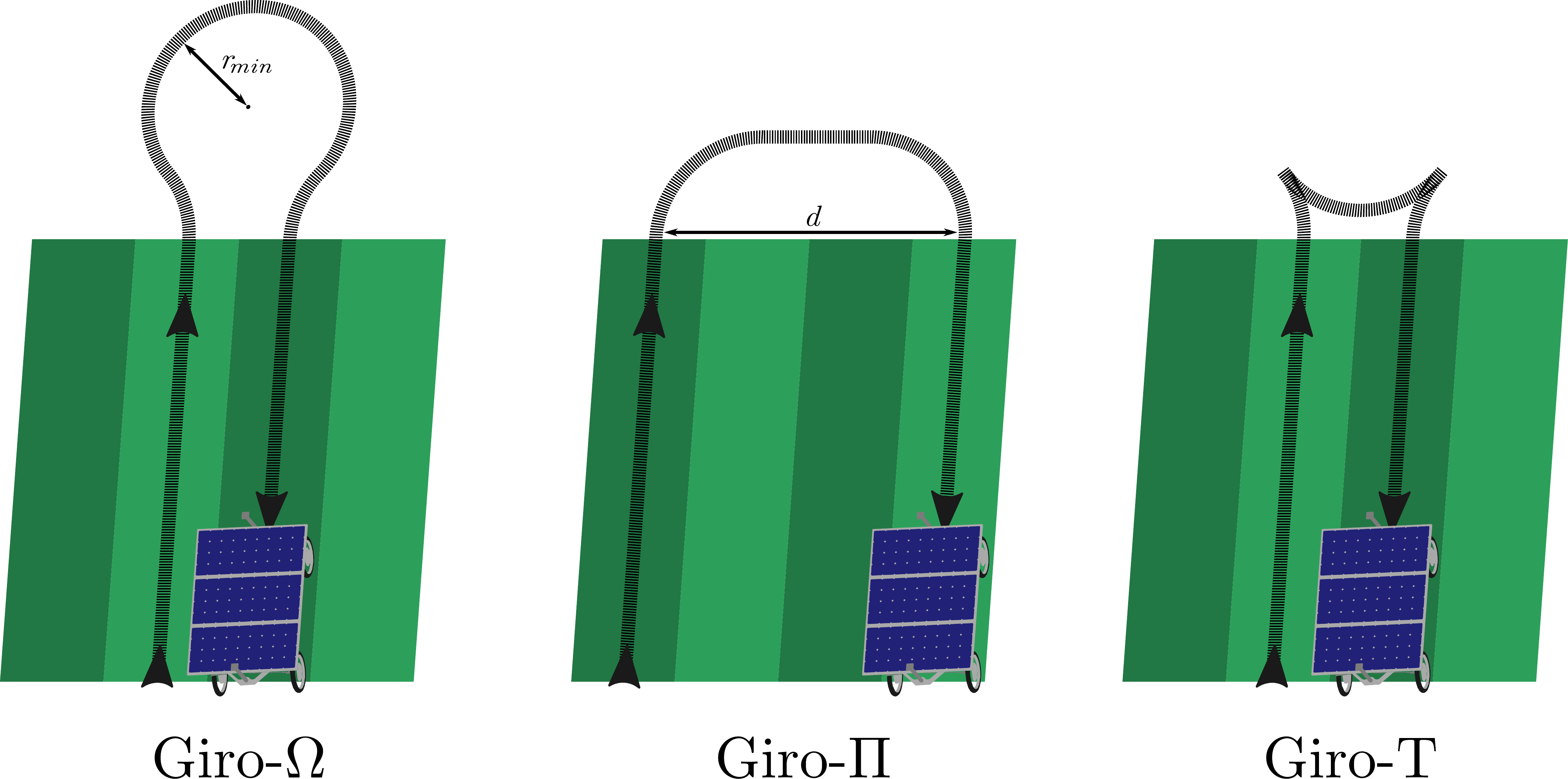} 
    \caption{Maniobras de giro entre franjas.}
    \label{fig:turning_types}
\end{figure}

 La longitud del trayecto que debe hacer un robot para pasar de una franja a otra depende del ángulo de giro del robot y obviamente de la distancia entre las mismas.
 A lo largo de la literatura, se destacan tres maniobras de giro diferentes como se pueden apreciar en la Figura~\ref{fig:turning_types}.
 Estos son el giro bucle (giro-$\Omega$), el giro de esquinas redondeadas (giro-$\Pi$) y el giro con reversa (giro-T).
 
 La ejecución del giro-$\Pi$ está restringida a $d \geq 2r_{min}$, donde $d$ es la distancia entre los centros de las franjas involucradas y $r_{min}$ es el radio mínimo de giro del robot.
 
 Para ejecutar el giro-T, el robot debe poder moverse en reversa. Si bien este giro implica menor recorrido que el \mbox{giro-$\Omega$}, el robot debe reducir su velocidad a cero dos veces para cambiar su sentido de circulación, lo cual podría resultar en una maniobra que implique mayor inversión de tiempo.

 Definamos ${F = \{f_1, \dots, f_n\}}$ como el conjunto de todas las franjas del campo a cubrir.
 La distancia entre los centros de dos franjas $f_i$ y $f_j$ la notaremos con $d_{ij}$.

 El giro-$\Pi$ produce el trayecto de menor longitud de los tres tipos, pero solo puede ejecutarse si ${d \geq 2r_{min}}$.
 Para el caso que ${d < 2r_{min}}$ se debe optar por el giro-$\Omega$ o el giro-T.
 Las siguientes ecuaciones definen las longitudes de las distintas maniobras.
\begin{align*}
    \Omega(d) &= r_{min}\bigg(3\pi-4\sin^{-1}\bigg(\frac{2r_{min} + d}{4r_{min}}\bigg)\bigg),\\
    \Pi(d) &= d + (\pi - 2)r_{min},\\
    \mathrm{T}(d) &= r_{min}\bigg(2\pi + \cos^{-1}\bigg(\frac{d + 2}{4r_{min}}\bigg)\bigg).
\end{align*}

 Definimos ${p : F \rightarrow \{1, \dots, n\}}$ como la función que retorna el orden en el cual una franja es recorrida.
 La función inversa ${p^{-1} : \{1, \dots, n\} \rightarrow F}$ indica qué franja debe recorrerse en cada paso.
 La secuencia de recorrido está dada entonces por ${\sigma = \langle p^{-1}(1), \dots, p^{-1}(n) \rangle}$

 Sea $L(f_i, f_j)$ la longitud mínima que debe recorrer el robot para pasar de la franja $f_i$ a la $f_j$. 
 Se pueden calcular los valores de dicha función para todo par de franjas, conociendo la distancia entre las franjas $d_{ij}$ y el ángulo mínimo de giro del robot $r_{min}$, y aplicando las formulas detalladas previamente.
 
 Nuestro objetivo resulta en encontrar la secuencia $\sigma$ que minimice la trayectoria total de los giros dada por
\begin{equation*}
    J(\sigma) = \sum_{i=1}^{n-1} {L(p^{-1}(i), p^{-1}(i+1))}.
\end{equation*}
 Para resolver este problema, podemos usar una representación de grafo $G = \{F,A\}$ donde el conjunto de franjas $F$ corresponde con los nodos del grafo, y las transiciones entre franjas $A = F\times F$ representan las aristas ponderadas.
 Cada arista $A_{ij}$ tendrá un peso o costo ${c_{ij} = L(f_i, f_j)}$, que determina la distancia mínima del trayecto para pasar de la franja $f_i$ a la franja $f_j$.
 De esta manera, el problema de encontrar la secuencia óptima de franjas, equivale a encontrar en el grafo el recorrido de menor costo que pase exactamente una vez por cada nodo.

 Este problema es equivalente al conocido problema del viajante (TSP por sus siglas en inglés \textit{Travelling Salesman Problem}).
 El TSP es un problema de tipo NP-difícil, lo que indica que el costo computacional requerido para resolverlo aumenta drásticamente cuando la dimensión del problema se incrementa.
 De todas formas, diversas heurísticas han sido desarrolladas para encontrar una solución cercana a la óptima en problemas de dimensiones relativamente grandes \cite{laporte1992thetravelling}.
\begin{figure}[!htbp]
    \centering
    \includegraphics[width=1\columnwidth]{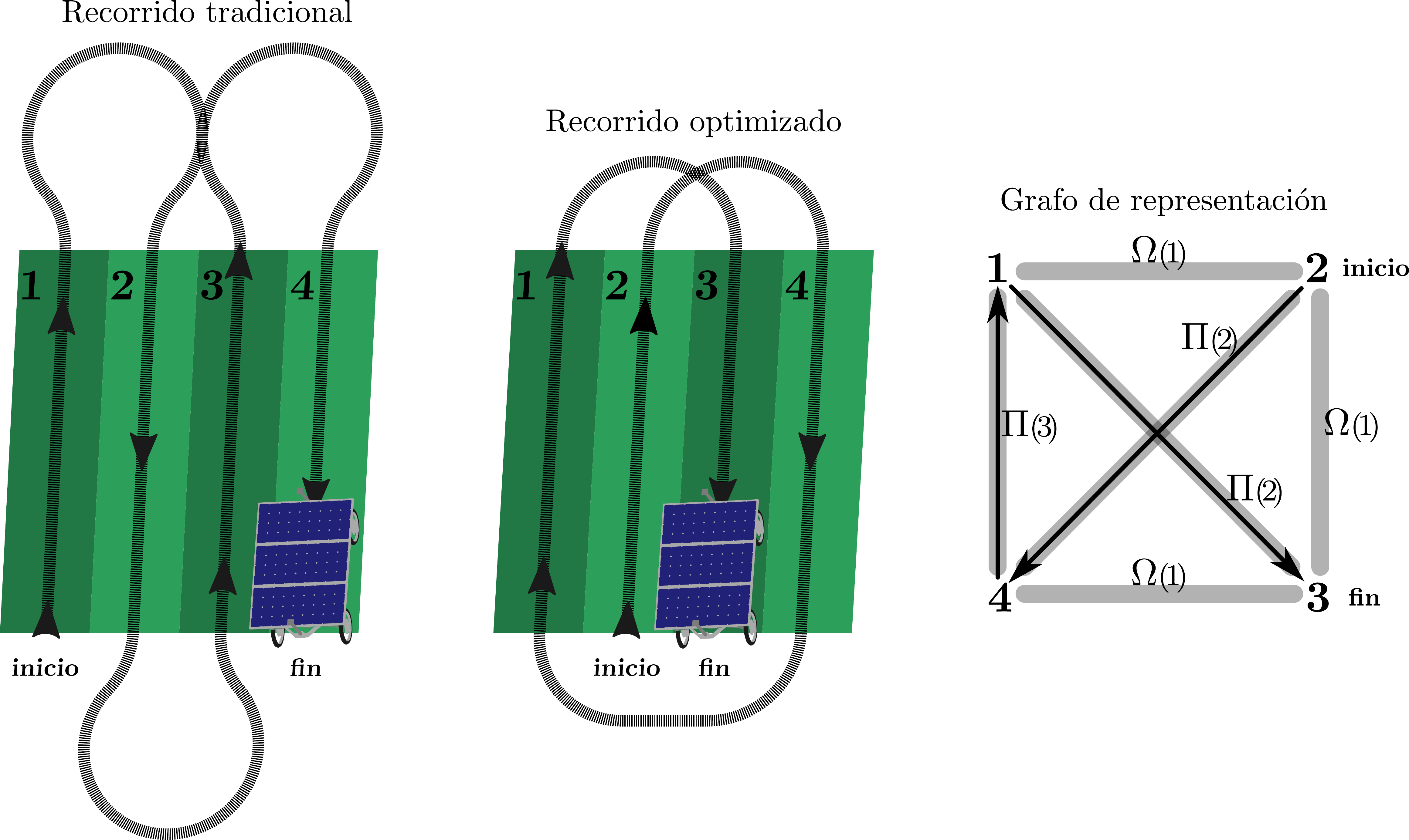} 
    \caption{Comparación recorrido tradicional y optimizado.}
    \label{fig:graph}
\end{figure}

 La Figura~\ref{fig:graph} muestra un ejemplo de grafo representando un campo con cuatro franjas y radio de giro igual al ancho de las franjas.
 Si se logra que el recorrido no posea ningún giro-$\Omega$, se puede reducir el ancho de las cabeceras utilizadas para girar y emplear más espacio del campo de manera productiva.

%% file: src/proposal.tex
 En esta sección discutiremos una posible mejora al procedimiento tradicional para la cobertura de espacios complejos con obstáculos.
 
 Analicemos el ejemplo del campo representado en la Figura~\ref{fig:decomposition}, con su descomposición en celdas ya definida. 
 El siguiente paso para obtener un plan de cobertura que abarque la totalidad del espacio, consiste en determinar un orden de recorrido de las celdas.
 Tomemos como ejemplo esta secuencia: ${\nu = \langle \cell{1},\cell{3},\cell{4},\cell{6},\cell{5},\cell{3},\cell{2} \rangle}$. 
 Notar que la celda \cell{3} inevitablemente debe ser visitada dos veces en el recorrido, pero solo en una de sus visitas el robot debe aplicar el patrón de cobertura generado para dicha celda.
 
 Para el último paso, consideramos la aplicación del método de elección de franjas descripto en \ref{sec:track_selection}, en cada una de las celdas de manera independiente. 
 Si no tenemos en cuenta la posición inicial y final del robot cuando generamos los patrones de cobertura para cada celda, al finalizar el recorrido de una de ellas el robot puede quedar en un punto relativamente alejado de la siguiente celda a recorrer en la secuencia $\nu$. 
 Si esto ocurre, el robot debe trasladarse desde el punto final de recorrido de una celda, al punto inicial del recorrido de la próxima, con el agregado de que debe hacerlo evitando pisar los cultivos.
 Esto es una desventaja si se busca minimizar la distancia total recorrida por el robot.
 
 Para obtener un plan de cobertura óptimo a nivel global, no podemos analizar la generación de las coberturas de cada celda de manera aislada del resto. 
 Una primera aproximación para mejorar el plan, surge al incorporar restricciones que determinen la posición inicial y final del robot al momento de abordar la cobertura de cada celda. 
 En la elección del orden de franjas, se agrega la restricción de que la franja inicial sea la más cercana al borde de la celda anteriormente recorrida y la franja final sea la más cercana a la próxima celda a recorrer. 
 Sin embargo, ya hemos visto por el análisis de los ángulos de giro del robot, que utilizar transiciones entre dos franjas contiguas generalmente no conduce al recorrido óptimo.
 
 Otro enfoque, podría ser intentar aplicar el método de elección de franjas directamente de forma global, considerando todas las franjas del campo al mismo tiempo.
 Sin embargo, esto puede ser no factible para ciertos espacios complejos.
 El método visto en la subsección \ref{sec:track_selection} se basa en que luego de haber recorrido una franja, el robot pueda pasar a cualquiera de las demás franjas del espacio considerado. 
 No tiene en cuenta en cuál extremo de la franja se encuentra el robot al momento de tener que elegir la siguiente. 
 Es por esto, que basta con un único nodo para representar cada franja en el grafo $G$. 
 Para que el método sea aplicable, ambos extremos de cada una de las franjas deben estar de alguna manera ``conectados'' con el resto de las franjas.
 
 En un campo rectangular ordinario (sin obstáculos) el requerimiento anterior se satisface, ya que todas las franjas se extienden de forma continua desde el borde inferior al borde superior. 
 A estos bordes superiores e inferiores, en los cuales el robot (o la maquinaria agrícola en general) puede realizar sus giros y circular para trasladarse de una franja a otra, se los conoce como cabeceras del campo. 
 En pocas palabras, para aplicar el método visto en \ref{sec:track_selection} todas las franjas en consideración deben compartir las cabeceras en ambos de sus extremos.
 
 En el campo de la Figura~\ref{fig:decomposition}, no se puede aplicar directamente y de manera global este método, ya que por ejemplo los extremos superiores de las franjas contenidas en la celda \cell{5} no comparten cabecera con franjas de ninguna otra celda. 
 Cuando el robot finalice el recorrido en el extremo superior de una de estas franjas, solo podrá elegir otra franja de la misma celda como la siguiente a recorrer. 
 No obstante, notemos que si restringimos el espacio de aplicación de la técnica a cada una de las celdas, el método resulta viable.
 
 En \cite{bochtis2009combined} se plantea de manera introductoria la posibilidad de aplicar el método de elección de franjas a nivel global en espacios complejos.
 Para lograr esto, se deben realizar ciertas modificaciones a la representación del grafo $G$ que formalizamos a continuación.

\begin{figure}[!htbp]
    \centering
    \includegraphics[width=0.7\columnwidth]{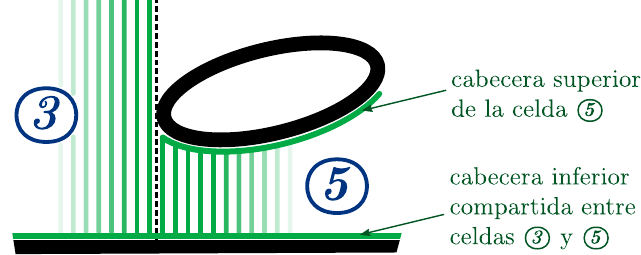} 
    \caption{Cabeceras de celdas.}
    \label{fig:tracks}
\end{figure}

 Notamos con ${G' = \{F', A'\}}$ al nuevo grafo, donde el conjunto de nodos ${F' = \{f'_{1}, f''_{1}, \dots, f'_{n}, f''_{n}\}}$ representa con cada $f'_{i}$ y $f''_{i}$ el extremo superior e inferior respectivamente de cada franja $f_{i}$ para $i \in \{1,\dots,n\}$.
 Por lo tanto, las aristas ${A' = F' \times F'}$ representan transiciones entre extremos de franjas y sus costos están determinados de la siguiente manera:
\begin{itemize}
    \item Costo cero entre $f'_i$ y $f''_i$ para indicar que cuando se ingresa a una franja por un extremo, se la recorra completamente hasta su otro extremo.
    \item Costo infinito entre un extremo superior y uno inferior (y viceversa) de cualquier par de franjas $f'_i$ y $f''_j$ con $i \neq j$, ya que dos extremos opuestos de franjas no podrían estar conectados por una misma cabecera.
    \item También costo infinito a dos extremos superiores (o inferiores) de franjas $f'_i$ y $f'_j$ con $i \neq j$, si no están conectados por una cabecera.
    \item Costos ordinarios (como los ya explicados en \ref{sec:track_selection}) dependientes del tipo de giro para dos extremos superiores (o inferiores) que sí estén conectados por una cabecera.
\end{itemize}
 La descomposición en celdas nos proporciona la información necesaria para saber cuál de los dos últimos grupos utilizar para determinar el costo de una transición entre dos extremos (no triviales).
 Cada celda define un par de cabeceras en los límites de las franjas contenidas.
 La Figura~\ref{fig:tracks} muestra un acercamiento del campo de ejemplo, donde se puede apreciar que la cabecera superior de la celda \cell{5} no está conectada con ninguna otra cabecera. 
 En cambio, la cabecera inferior de \cell{5} se encuentra conectada con la cabecera inferior de la celda \cell{3}. 
 En consecuencia, al finalizar el recorrido de una franja de \cell{5} en el extremo superior, solamente se puede elegir otra franja de la misma celda como la siguiente a recorrer. 
 En caso contrario, si se finaliza en el extremo inferior de \cell{5}, sí se puede elegir una franja de otra celda, como ser la \cell{3}.
 Este método amplía las posibles soluciones consideradas a nivel global, ya que el plan de cobertura puede entrar y salir varias veces de una misma celda, posibilitando la generación de una cobertura más eficiente. 
 Recordar que anteriormente mencionamos que la celda \cell{3} debe ser visitada al menos dos veces.
 Con este método, no es necesario que se cubra completamente la celda en una única de las visitas, sino que se puede realizar una cobertura parcial en cada una.
 
 Notar que la solución tradicional, en la cual se generan coberturas completas de cada celda antes de pasar a la siguiente, está incluida dentro de las posibles soluciones consideradas por este método, lo que asegura la obtención de una solución no menos eficiente. 
 Sin embargo, cabe destacar que la complejidad del algoritmo es mayor debido al aumento en la dimensión del grafo, ya que se consideran todas las franjas a la vez y cada una es representada por dos nodos en lugar de uno.

%% file: src/conclusions.tex
En este trabajo se realiza una revisión general de los métodos de planificación de cobertura, poniendo especial foco en aquellos aplicables a entornos agrícolas.
Se presentan formalmente las técnicas de descomposición en celdas y elección de orden óptimo de franjas.
Se discute la forma tradicional de generación de un plan de cobertura global en espacios complejos con obstáculos y se analiza una posible mejora combinando las técnicas presentadas. Como trabajo futuro, se plantea la valoración cuantitativa de las mejoras.

La revisión y análisis de los métodos de cobertura aquí expuestos establecen una base sólida para la selección e implementación de los mismo en el módulo de planeamiento de trayectorias del robot desmalezador desarrollado en el instituto CIFASIS (CONICET-UNR).
Actualmente nos encontramos realizando la implementación de los algoritmos descriptos. 
Los mismos serán evaluados en primer lugar en el entorno de simulación \textit{Gazebo}, para posteriormente utilizarlos en el prototipo real.
%
\begin{figure}[!htbp]
    \centering
    \includegraphics[width=0.65\columnwidth]{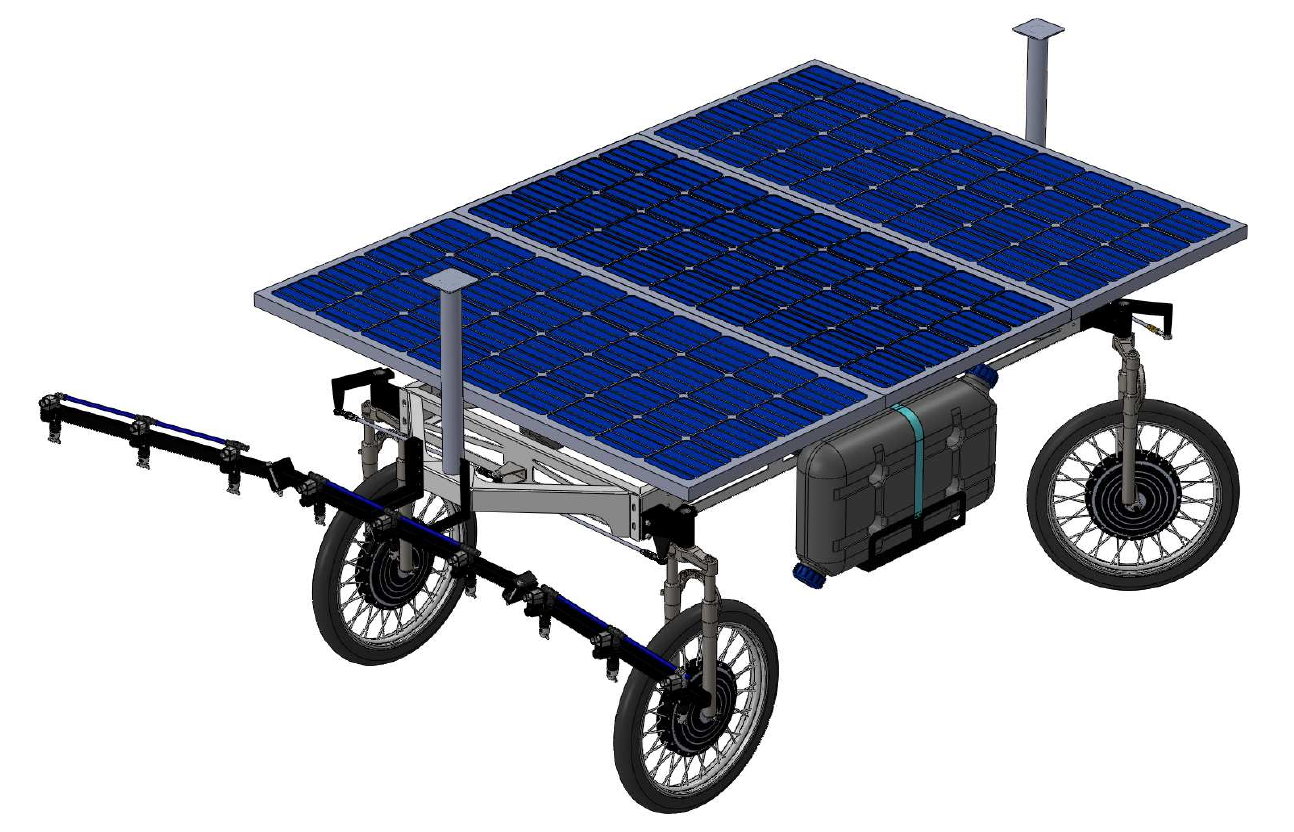} 
    \caption{Modelo del robot desmalezador en \textit{Solidworks}.}
    \label{fig:solidworks}
\end{figure}

%% file: main.bbl
\begin{thebibliography}{10}
\providecommand{\url}[1]{#1}
\csname url@samestyle\endcsname
\providecommand{\newblock}{\relax}
\providecommand{\bibinfo}[2]{#2}
\providecommand{\BIBentrySTDinterwordspacing}{\spaceskip=0pt\relax}
\providecommand{\BIBentryALTinterwordstretchfactor}{4}
\providecommand{\BIBentryALTinterwordspacing}{\spaceskip=\fontdimen2\font plus
\BIBentryALTinterwordstretchfactor\fontdimen3\font minus
  \fontdimen4\font\relax}
\providecommand{\BIBforeignlanguage}[2]{{%
\expandafter\ifx\csname l@#1\endcsname\relax
\typeout{** WARNING: IEEEtran.bst: No hyphenation pattern has been}%
\typeout{** loaded for the language `#1'. Using the pattern for}%
\typeout{** the default language instead.}%
\else
\language=\csname l@#1\endcsname
\fi
#2}}
\providecommand{\BIBdecl}{\relax}
\BIBdecl

\bibitem{vougioukas2019agricultural}
S.~Vougioukas, ``{Agricultural Robotics},'' \emph{Annual Review of Control,
  Robotics, and Autonomous Systems}, vol.~2, pp. 365 -- 392, 2019.

\bibitem{choset1998coverage}
H.~Choset and P.~Pignon, ``{Coverage Path Planning: The Boustrophedon Cellular
  Decomposition},'' in \emph{Field and Service Robotics (FSR)}, 1998, pp. 203
  -- 209.

\bibitem{decarvalho1997complete}
R.~N. {De Carvalho}, H.~A. Vidal, P.~Vieira, and M.~I. Ribeiro, ``{Complete
  coverage path planning and guidance for cleaning robots},'' in
  \emph{Proceeding of the IEEE International Symposium on Industrial
  Electronics}, vol.~2, 1997, pp. 677 -- 682.

\bibitem{dakulovic2011complete}
M.~Dakulović, S.~Horvatić, and I.~Petrović,
  ``\href{https://doi.org/10.3182/20110828-6-IT-1002.03400}{Complete Coverage
  D* Algorithm for Path Planning of a Floor-Cleaning Mobile Robot},''
  \emph{IFAC Proceedings Volumes}, vol.~44, pp. 5950 -- 5955, 2011.

\bibitem{acar2003path}
E.~Acar, H.~Choset, Y.~Zhang, and M.~Schervish, ``{Path Planning for Robotic
  Demining: Robust Sensor-Based Coverage of Unstructured Environments and
  Probabilistic Methods},'' \emph{The International Journal of Robotics
  Research}, vol.~22, pp. 441 -- 466, 2003.

\bibitem{acar1999sensor}
E.~Acar, M.~Simmons, M.~Rosenblatt, M.~Roth, M.~Berna, Y.~Mittlefehldt, and
  H.~Choset, ``{Sensor Based Coverage of Unknown Environments for Land Mine
  Detection},'' in \emph{Sixteenth National Conference on Artificial
  Intelligence}, 1999, pp. 932 -- 933.

\bibitem{yang2004aneural}
S.~X. {Yang} and C.~{Luo}, ``{A neural network approach to complete coverage
  path planning},'' \emph{IEEE Transactions on Systems, Man, and Cybernetics},
  vol.~34, pp. 718 -- 724, 2004.

\bibitem{farsi1994robot}
M.~A. Farsi, K.~Ratcliff, J.~P. Johnson, C.~R. Allen, K.~Z. Karam, and
  R.~Pawson, ``{Robot control system for window cleaning},'' \emph{Proceedings
  of American Control Conference}, vol.~1, pp. 994 -- 995, 1994.

\bibitem{pire2019rosario}
T.~Pire, M.~Mujica, J.~Civera, and E.~Kofman,
  ``\href{https://doi.org/10.1177/0278364919841437}{The Rosario dataset:
  Multisensor data for localization and mapping in agricultural
  environments},'' \emph{Intl. J. of Robotics Research}, vol.~38, pp. 633--641,
  2019.

\bibitem{choset2001coverage}
H.~Choset, ``{Coverage for robotics – A survey of recent results},''
  \emph{Annals of Mathematics and Artificial Intelligence}, no.~31, pp. 113 --
  126, 2001.

\bibitem{galcernan2013asurvey}
E.~Galceran and M.~Carreras,
  ``\href{https://doi.org/10.1016/j.robot.2013.09.004}{A survey on coverage
  path planning for robotics},'' \emph{Journal of Robotics and Autonomous
  Systems}, vol.~61, pp. 1258 -- 1276, 2013.

\bibitem{khan2017oncomplete}
A.~Khan, I.~Noreen, and Z.~Habib, ``{On Complete Coverage Path Planning
  Algorithms for Non-holonomic Mobile Robots: Survey and Challenges},''
  \emph{Journal of Information Science and Engineering}, vol.~33, pp. 101 --
  121, 2017.

\bibitem{bochtis2008minimising}
D.~Bochtis and S.~Vougioukas, ``{Minimising the non-working distance travelled
  by machines operating in a headland field pattern},'' \emph{Biosystems
  Engineering}, vol. 101, pp. 1 -- 12, 2008.

\bibitem{hammed2011driving}
I.~Hameed, D.~Bochtis, and C.~Sørensen, ``{Driving Angle and Track Sequence
  Optimization for Operational Path Planning Using Genetic Algorithms},''
  \emph{Applied Engineering in Agriculture}, vol.~27, pp. 1077--1086, 2011.

\bibitem{oksanen2009coverage}
T.~Oksanen and A.~Visala, ``{Coverage path planning algorithms for agricultural
  field machines},'' \emph{Journal of Field Robotics}, vol.~26, pp. 651 -- 668,
  2009.

\bibitem{milnor1969morse}
J.~Milnor, M.~Spivak, and R.~Wells, \emph{{Morse Theory}}.\hskip 1em plus 0.5em
  minus 0.4em\relax Princeton University Press, 1969, vol.~51.

\bibitem{acar2002morse}
E.~Acar, H.~Choset, A.~Rizzi, P.~Atkar, and D.~Hull, ``{Morse Decompositions
  for Coverage Tasks},'' \emph{Intl. J. of Robotics Research}, vol.~21, pp. 331
  -- 344, 2002.

\bibitem{clarke1990optimization}
F.~Clarke, \emph{{Optimization and Nonsmooth Analysis}}.\hskip 1em plus 0.5em
  minus 0.4em\relax Society for Industrial and Applied Mathematics, 1990.

\bibitem{laporte1992thetravelling}
G.~Laporte, ``{The traveling salesman problem: An overview of exact and
  approximate algorithms},'' \emph{European Journal of Operational Research},
  vol.~59, no.~2, pp. 231 -- 247, 1992.

\bibitem{bochtis2009combined}
D.~Bochtis and T.~Oksanen, ``{Combined coverage and path planning for field
  operations},'' in \emph{Precision agriculture}, 2009, pp. 521 -- 527.

\end{thebibliography}
